\documentclass[lettersize,journal]{IEEEtran}
\usepackage{etoolbox}
\pdfoutput=1
\makeatletter
\patchcmd{\@makecaption}
  {\scshape}
  {}
  {}
  {}
\makeatletter
\patchcmd{\@makecaption}
  {\\}
  {.\ }
  {}
  {}
\usepackage{amsmath,amsfonts}
\usepackage{algorithmic}
\usepackage{algorithm}
\usepackage{array}
\usepackage[caption=false,font=normalsize,labelfont=sf,textfont=sf]{subfig}
\usepackage{textcomp}
\usepackage{stfloats}
\usepackage{url}

\usepackage{verbatim}
\usepackage{graphicx}
\usepackage{cite}
\begin{document}

\title{TRUST: An Accurate and End-to-End Table structure Recognizer Using Splitting-based Transformers}

\author{Zengyuan guo, Yuechen Yu, Pengyuan Lv, Chengquan Zhang, Haojie Li, Zhihui Wang, Kun Yao, Jingtuo Liu, Jingdong Wang

\thanks{This paper was produced by the Baidu Inc.}
\thanks{Haojie Li and Zhihui Wang are professors of Dalian University of technology}}



\maketitle

\begin{abstract}
Table structure recognition is a crucial part of document image analysis domain~\cite{hashmi2021current}. Its difficulty lies in the need to parse the physical coordinates and logical indices of each cell at the same time. However, the existing methods are difficult to achieve both these goals, especially when the table splitting lines are blurred or tilted. In this paper, we propose an accurate and end-to-end transformer-based table structure recognition method, referred to as TRUST. Transformers are suitable for table structure recognition because of their global computations, perfect memory, and parallel computation. By introducing novel Transformer-based Query-based Splitting Module and Vertex-based Merging Module, the table structure recognition problem is decoupled into two joint optimization sub-tasks: \emph{multi-oriented table row/column splitting} and \emph{table grid merging}. The Query-based Splitting Module learns strong context information from long dependencies via Transformer networks~\cite{vaswani2017attention}, accurately predicts the multi-oriented table row/column separators, and obtains the basic grids of the table accordingly. The Vertex-based Merging Module is capable of aggregating local contextual information between adjacent basic grids, providing the ability to merge basic girds that belong to the same spanning cell accurately. We conduct experiments on several popular benchmarks including PubTabNet\cite{zhong2020image} and SynthTable\cite{qasim2019rethinking}, our method achieves new state-of-the-art results. In particular, TRUST runs at 10 FPS on PubTabNet, surpassing the previous methods by a large margin.
\end{abstract}


\section{Introduction}
\IEEEPARstart{T}{able} Structure Recognition aims to recognize the internal structure of a table. It is a fundamental task in document understanding and has numerous practical applications~\cite{zanibbi2004survey}, such as question answering, dialogue systems, table-to-text, \emph{etc}. With the increasing number of documents containing tables, automated reading of tables within these images has become an urgent task.

\begin{figure}[t]
\begin{center}
\includegraphics[width=1.0\linewidth]{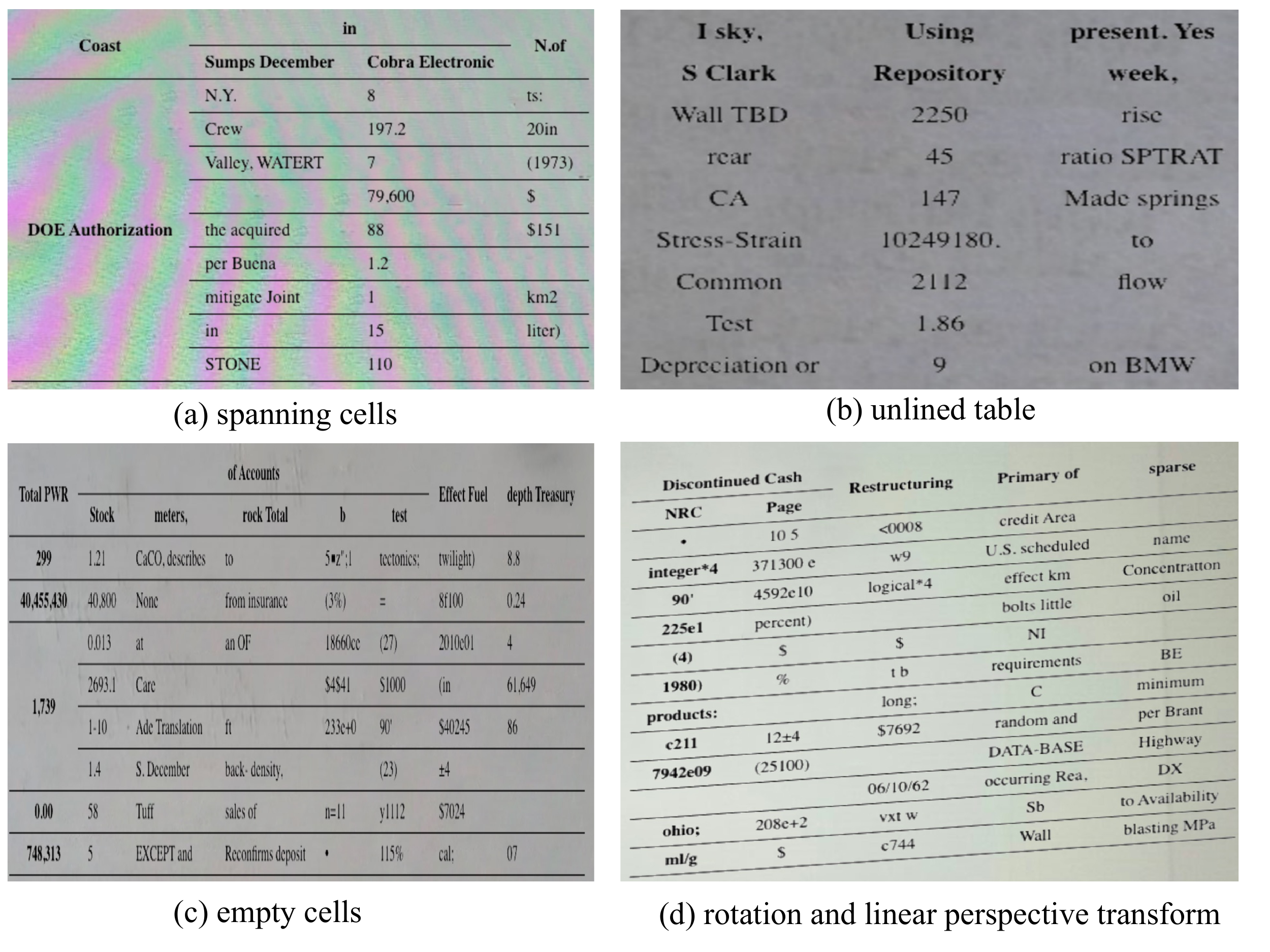}
\end{center}
\caption{Overview of different types of table structure recognition methods.}
\label{fig:hard_sample}
\end{figure}

Through studied for years, table structure recognition is still a very open research problem. The main difficulty lies in the need to parse the exact bounding box and logical index of each cell at the same time. In particular, four types of degradation and variations cause various problems in most current table structure recognition systems, as illustrated in Fig.\ref{fig:hard_sample}. First, spanning cells that occupy at least two rows or columns are more important than other simple cells on tables because spanning cells are more likely to be table headers in a table\cite{chi2019complicated}. Second, parsing unlined tables or partially lined tables is more difficult than lined tables, because there are no explicit visual cues that delimit cells, columns, and rows. Third, empty cells are easier to neglect and more difficult to be located than non-empty cells in tables. Fourth, rotation and linear perspective transformation may degrade strongly the performance of table structure recognition.

Recent efforts have been devoted to improving the performance of table structure recognition which can be summarized into three categories: (1) Component-based Approaches (2) Sequence-based Approaches (3) Splitting-based Approaches. Unfortunately, the Component-based approaches such as DeepDeSRT~\cite{schreiber2017deepdesrt}, TableNet~\cite{paliwal2019tablenet} and LGPMA~\cite{qiao2021lgpma} still suffer from boundary ambiguity problems in unlined tables and cannot achieve decent performance in complex scenarios such as tables with empty cells. Besides, the Sequence-based Approaches such as EDD~\cite{zhong2020image} strongly depend on a large amount of data for end-to-end training and the generalization will drop sharply when encountering unseen data. Moreover, they often fail to regress accurate cell boundaries. Significantly, the Splitting-based Approaches provide strong generative capabilities for different kinds of table images because they mainly focus on capturing global and local visual context in tables such as the row or column separators or the linking relationships between a pair of adjacent basic cells, which will not change very large among different kinds of table images. Also, the Splitting-based approaches can attain more accurate cell locations compared with Component-based approaches and Sequence-based approaches. Our proposed TRUST follows the Splitting-based Approaches. 

However, recent Splitting-based approaches such as SPLERGE~\cite{tensmeyer2019deep} and SEM~\cite{zhang2022split} may suffer from following disadvantages: (1) The pipeline of SEM is inefficient, which may involve time-consuming Region of Interest (RoI)~\cite{ren2015faster} operations and context features extraction via BERT~\cite{DBLP:journals/corr/abs-1810-04805}. (2) SPLERGE trained two isolate split-model and merge-model which may increase the difficulty of optimization compared with training in an end-to-end fashion. (3) Existing Splitting-based approaches can not handle well tables with rotation and linear perspective transform.

In this paper, we propose an end-to-end Transformer-based table structure recognition method. Our method addresses the challenges in table structure recognition via an innovative encoder-decoder architecture as illustrated in Fig.\ref{fig:overview}. The Convolutional Neural Networks with FPN are used as the backbone feature extractor. We enable table structure recognition with a Query-based Splitting Module, which introduces angle classification and starting point prediction for multi-oriented row/column separators. Through these predicted separators, a fine grid structure of the table is generated. 

In addition, we design a novel Vertex-based Merging Module, to calculate features of all intersection of row separators and column separators, a.k.a vertices. With these features of vertices, a self-attention mechanism is built to scan all vertices and predicts which basic grid pairs should be merged in four directions including \emph{(top-left, top-right), (top-right, down-right), (down-left, down-right) and (top-left, down-left)} around vertices. Vertex-based Merging Module helps to merge adjacent grids together to recover the spanning table cells more accurately, regardless of unlined tables or tables with empty cells. Our model is trained in an end-to-end fashion and the results show the effectiveness of our method.

The major contributions of this work can be summarized in the following three points:
\begin{enumerate}
\item We present a novel end-to-end framework named TRUST to tackle the tasks of table structure recognition, which leverages multi-headed self- and cross-attention mechanisms between the visual feature maps and row/column features to capture contextual information from long dependencies more efficiently and effectively. Furthermore, we design a novel Query-based Splitting Module and Vertex-based Merging Module to extract semantic features of the row/column separators and vertices, leading to more accurate table structure recognition in a split-merge manner.
\item Our Splitting-based TRUST can handle well most categories of tables, including those that are unlined or partially lined, and those with empty cells or spanning cells. Moreover, TRUST can recognize the structure of rotating tables, which is not solved very well by the previous Splitting-based methods.
\item We develop an end-to-end trainable table structure recognition method that demonstrates superior performance over some public datasets including the PubTabNet and SynthTable.
\end{enumerate}

\section{Related Works}
Quite a number of table recognition techniques have been reported in recent years~\cite{scarselli2008graph, xue2019res2tim, siddiqui2019deeptabstr}, and most of them can be broadly classified into three categories. The first one follows a bottom-up approach which first detects text parts or basic cell parts and then links them up to form a table structure through graph neural networks or post-processing. The second follows a sequence decoding framework which treats table recognition as a image-to-sequence problem. The third follows a split-merge approach which obtains the basic table grids through dense splitting lines prediction, and then merge some of them to form spanning cells. 

\textbf{Component-based Approaches}. Many conventional methods follow a bottom-up approach that first detects text or basic cell parts and then connects them to form a table structure. Popular table structure recognition methods include DeepDeSRT\cite{schreiber2017deepdesrt}, ReS2Tim\cite{xue2019res2tim}, DeepTabStR\cite{siddiqui2019deeptabstr}, etc. More recent methods explore Graph Neural Networks to link the basic components. For example, TIES~\cite{qasim2019rethinking} combines CNN~\cite{krizhevsky2012imagenet} and GNN~\cite{scarselli2008graph} to construct a bottom-up model to recognize the table structure. TabStruct-Net\cite{raja2020table} first detects individual cells and then links them to get table structure by graphs. Similarly, in NCGM\cite{liu2021neural}, it leverages graphs and modality interaction to boost the multi-modal representation for complex scenarios. Though Component-based Approaches are efficient, the big challenge is that these methods often fail to detect spanning cells and require extra cell detection networks which reduce the efficiency.

\textbf{Sequence-based Approaches}. Methods that directly reconstruct table structure in image-to-sequence manners become popular recently as reconstructing table structure in one shot avoids the extra linking process. EDD\cite{zhong2020image} utilizes an attention-based encoder-dual-decoder architecture to convert images of tables into HTML code. Its structure decoder reconstructs the table structure and directly recognizes cell content by the cell decoder at the same time. Though direct recognition of table structure is efficient, the big challenge is that these methods depend largely on the amount of trainable data and often fail to regress accurate cell locations.

\textbf{Splitting-based Approaches}. Splitting-based methods divide table structure recognition into two phases. They split the table into basic grid elements in which adjacent ones are then merged to recover spanning cells. For example, SPLERGE\cite{tensmeyer2019deep} first predicts the basic table grid pattern using Row Projection Networks and Column Projection Networks with novel projection pooling and then combines them to get table structure. Similarly, in SEM\cite{zhang2022split}, a splitter is applied to obtain the fine grid structure of the table by predicting the potential regions of the table row/column separators. It also enhances the representational power of each table cell by modeling the textual information via transformer networks and merging these table cells through the attention mechanism.

Our proposed TRUST follows the splitting-based approaches. Different from existing techniques, we predict row/column separators using a transformer decoder namely a Query-based Splitting Module, which can more effectively and efficiently deals with unconstrained table. Additionally, a novel Vertex-based Merging Module in which the vertex's representation is efficiently constructed of the learnable row/column representations from Query-based Splitting Module is introduced to merge table grids. Compared with the training of two independent modules in SPLERGE\cite{tensmeyer2019deep}, the whole framework of TRUST can be trained in an end-to-end manner and achieve better performance.

\section{Proposed Method}
\subsection{Overview}
\begin{figure*}
\begin{center}
\includegraphics[width=1.0\linewidth]{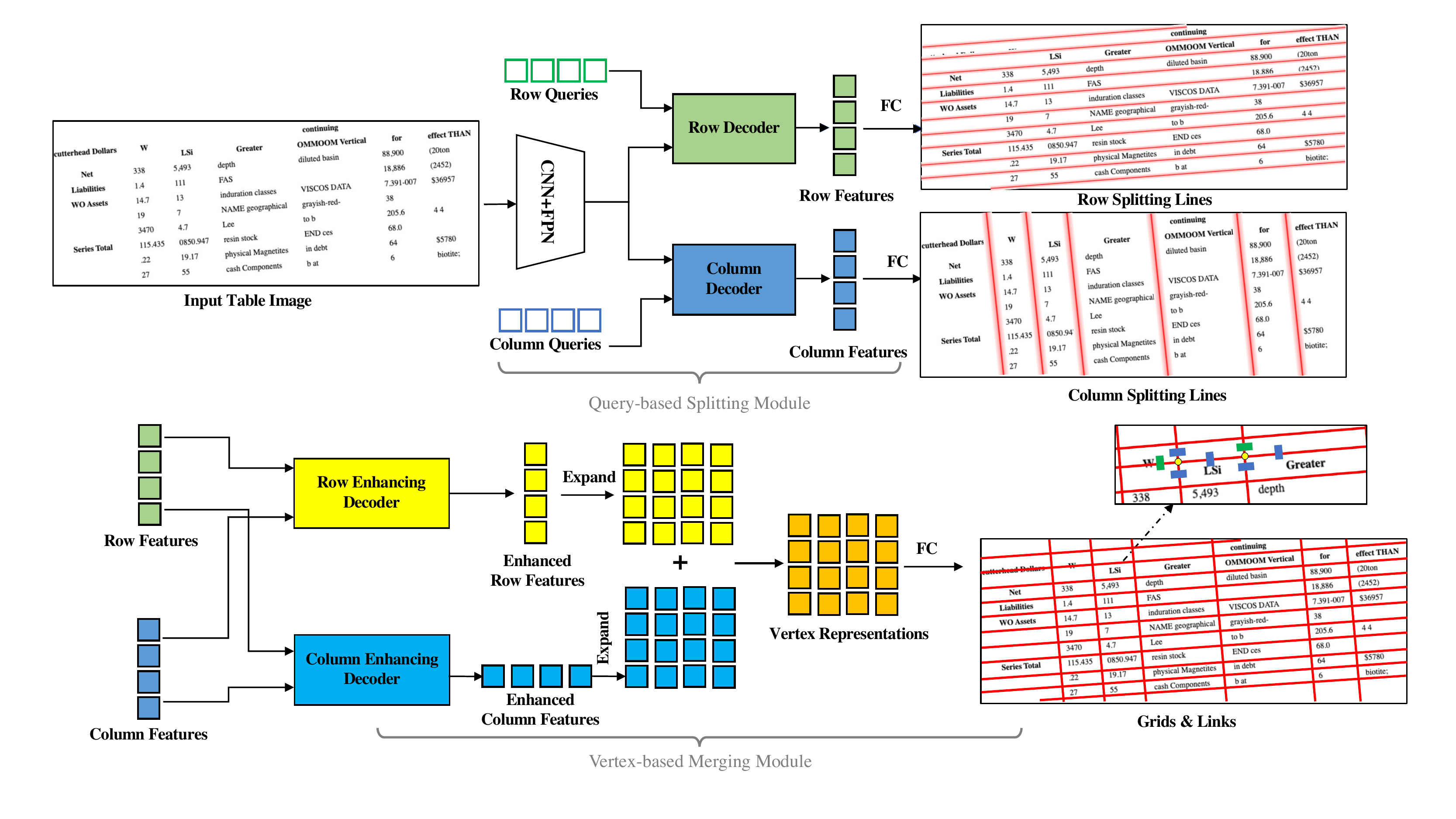}
\end{center}
\caption{An overview of the proposed TRUST. It consists of a CNN backbone, a Query-Based Splitting Module, and a Vertex-based Merging Module. The features of row/column separators are extracted and then generate row splitting lines and column splitting lines, forming a fine grid structure by the Query-Based Splitting Module. The row/column features are further fed into the Vertex-based Merging Module to predict the linking relations between adjacent basic cells}
\label{fig:overview}
\end{figure*}

We describe the details of our TRUST, As Shown in Fig.\ref{fig:overview}, it consists of three main components: a CNN backbone, a Query-Based Splitting Module, and a Vertex-based Merging Module. 

We use a ResNet18\cite{he2016deep} as the visual feature encoder of TRUST which computes increasingly high-level visual features as the layers become deeper. To alleviate the size problem of table and text, we adopt the FPN\cite{lin2017feature} strategy to merge features of different resolutions. Then in the Query-based Splitting Module, a transformer network is used as the feature decoder, in which visual features and learnable row/column position embedding features~\cite{carion2020end} are jointly used to capture features in a horizontal direction and vertical direction, respectively. We apply FFNs to the learnable row/column representations in the previous stage, generating the row and column separators of arbitrary orientations in the table. Through these predicted separators, a fine grid structure of the table is generated and each cell in this grid is a basic element of the table. Finally, those generated basic grids are further merged if they belong to the same spanning cells by a Vertex-based Merging Module. The feature representation of each vertex can be efficiently constructed by fusing the associated row and column features extracted by the Query-based Splitting Module. After further enhancing the vertex representation via FFNs, they are used to predict the merging results between adjacent basic grids.

\subsection{Query-Based Splitting Module}
As illustrated in Fig.\ref{fig:overview}, the proposed Query-Based Splitting Module takes visual features and row/column embedding features as inputs. In the Transformer decoder, visual features $F^{V} \in R^{H \times W \times d}$ obtained by the CNN encoder are firstly flattened to $R^{(H \times W) \times d} $ and fed to the Transformer decoder as the key and value of attention mechanism. In the meantime, the position indexes $(0,1,..,N-1)$ of rows and $(0,1,...,M-1)$ of columns are fed to Embedding layers to get embedding features $F^{embed} \in R^{N/M \times d} $, which are used as the queries of attention mechanism. $N$ and $M$ represent the predefined maximum number of horizontal and vertical separators in the table.

Following the row/column Transformer decoder, three fully connected layer produces the final prediction $(c_{i/j}, o_{i/j}, a_{i/j})$ for each row/column queries. $c_{i/j}$ means whether the row/column query is classified as a horizontal/vertical separator or not, $o_{i/j}$ means the offset value of each predicted horizontal/vertical separator intersecting the left/top boundary of the table, and $a_{i/j}$ means the predicted rotation angle of each row/column separator. Note that the offset value and rotation angle are only meaningful when the row/column queries are classified as positive. Based on the forecast results, a fine grid structure of the table can be generated, as shown in the bottom-right of Fig.\ref{fig:overview}

The proposed Query-Based Splitting Module addresses the unconstrained tables better from two aspects. First, the use of the self-attention mechanism in Transformers helps to capture contextual information from global long dependencies, which is very helpful for cases with blurred splitting lines and plentiful empty cells. Second, the prediction output of the separator with rich attributes can well describe the scene of tilted table lines.

\subsection{Vertex-based Merging Module}
We can accurately represent a simple merge-free table without spanning cells based only on the fine grid structure generated by Query-Based Splitting Module. However, when table contains spanning cells, the basic cells belonging to same spanning cell need to be merged. To solve this problem, we introduce Vertex-based Merging Module to model cell merging. 

First, the intersection of each horizontal and vertical separator represents a table vertex, whose features can be efficiently obtained by the fusion of horizontal and vertical separator features. We have previously obtained the horizontal separator feature $F^{r} \in R^{N \times d} $ and vertical separator feature $F^{c} \in R^{M \times d} $ in Query-Based Splitting Module. We then expand them into a feature with a shape of $N \times M \times d$ and add them together, getting the feature representation of $N \times M$ vertices, where N and M represent the number of horizontal and vertical separators. To improve the perception of vertex features to the context information of table rows and columns, before the horizontal and vertical separator features are fused, we perform cross-feature enhancement on them. The enhancement method is shown at the bottom of Fig.\ref{fig:overview}. When enhancing the horizontal separators, we use the horizontal separator feature $F^{r} \in R^{N \times d}$ as query, the vertical separator feature $F^{c} \in R^{M \times d}$ as key and value, and feed them into Transformer decoder layers to get the enhanced horizontal separator features. Using a similar operation, we can get the enhanced features of the vertical separators. The features of all vertices in the table are enhanced by the cross-attention mechanism between row and column separators. Each vertex will predict 4 attribute values, which are used to predict whether the 4 grid pairs around it should be merged. Specifically, 4 grid pairs include \emph{(top-left, top-right), (down-left, down-right), (top-left, down-left) and (top-right, down-right)}.

From the Query-based Splitting Module and Vertex-based Merging Module mentioned above, we can get the horizontal and vertical separators and vertices of the table, and further determine the information of the basic table grids and the merged information between the grids. From this information, we can form a variety of complex table structures.

\subsection{Ground Truth and Loss Function} 

In the above section, we refer to the two components of the table, the splitting lines and vertexes, and their respective attributes. In this section, we will describe how attribute labels are derived and the design of the Loss function.

\begin{figure}
\begin{center}
\includegraphics[width=1.0\linewidth]{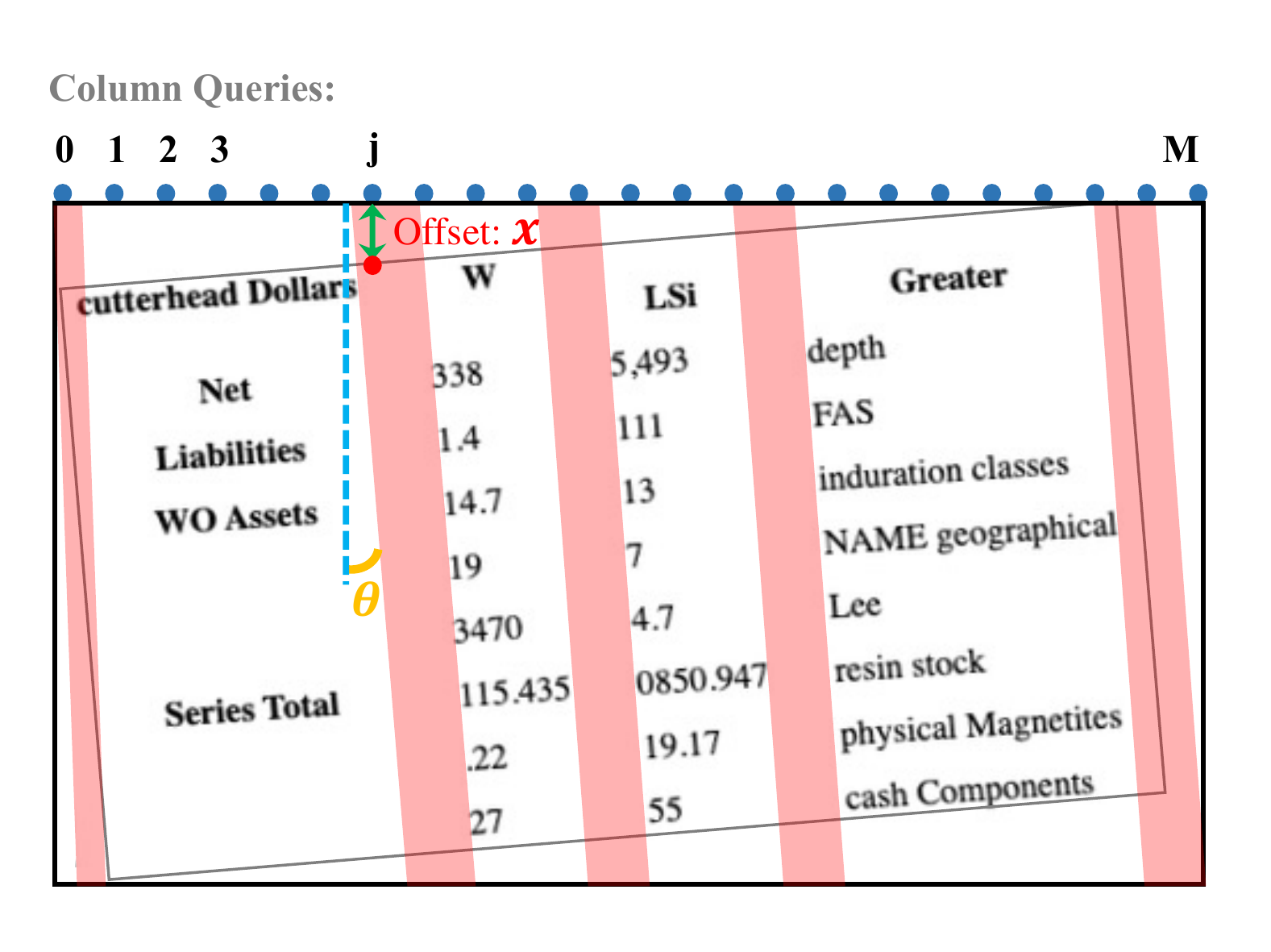}
\end{center}
\caption{The label generation process of columns: the j-th query is a positive column query; red regions represent column separators; red points represent the start point of positive column queries and $\theta$ represents the rotation angle of positive column queries. The label generation process of rows is similar.}
\label{fig:gt_generation}
\end{figure}

\textbf{Ground Truth for Query-based Splitting Module.} We expanded the definition of splitter in SPLERGE\cite{tensmeyer2019deep} to support inclined separators. As illustrated in Fig.\ref{fig:gt_generation}, we use parallelogram to express the column table separators. The format of parallelogram can maximize the area of the separator regions without intersecting non-spanning cell content,  and it is especially suitable for the inclined tables.

As described earlier, we predefined M queries for column table separators. In order to ease the difficulty of learning in query mode, we evenly distribute the column queries along the horizontal direction of the table image according to the index value of query. Therefore, the $j-th$ index corresponds to the $(j * w / M)$ horizontal position in the image, where the $w$ means the width of table image.

Next, we use this predefined position information to determine whether the query falls in the area of a column separator. If so, the category of query is set to positive class. At the same time, the angle label is set to $\theta$, which is the angle between the corresponding quadrilateral and the vertical direction. In addition, by drawing a vertical line with horizontal coordinate $(j * w / M)$ and the point intersecting the top boundary on the table is easily obtained, we can get the vertical offset value $x$. With the predefined horizontal position information and vertical offset value, plus the rotation angle, we can draw an accurate column splitting line. Using a similar operation method, we can also obtain the labels of n row split lines queries.

Once determining the positive horizontal and vertical queries of a table, the basic grids of the table are determined. 

\textbf{Ground Truth for Vertex-based Merging Module.} For partially complex tables, a portion of the text region may span more than one base cell, so some basic cells need to be merged. Merge labels are reflected in attributes of intersection points. The attributes of the intersection points have the following four dimensions, that is, the upper, lower, left and right, which represent the four merging proposals respectively. The attributes of the intersection points indicate whether the adjacent basic cells around the intersection point need to be merged or not. If two cells need to be merged, the attributes of their common intersection point should be set to positive. For example, two basic cells with index (i, j) and index (i + 1, j) need to be merged, i means row i and j means column j. Then the (top-left, bottom-left) attribute value of their common intersection point with index (i, j) is set to positive, and the (top-right, bottom-right) merged attribute value with index (i, j-1) is set to 1. The cell has the same index value as its bottom right corner point. 

\textbf{Loss Function.} Our model is trained in an end-to-end fashion, where the training loss is a weighted combination of multiple functions from Query-Based Splitting Module and Vertex-based Merging Module. Overall, the loss function is a weighted sum of the three losses:

\begin{equation}
\begin{aligned}
  & L(y_{row}, c_{row}, y_{col}, c_{col}, y_{ang}, c_{ang}, \hat{s}, s, y_{lnk}, c_{lnk})\\
  & = \frac{1}{N_{r}}L_{bce}(y_{row}, c_{row})  
  + \frac{1}{N_{c}}L_{bce}(y_{col}, c_{col})  \\
  & + \frac{1}{N_{pos}}L_{ce}(y_{ang}, c_{ang}) 
  + \frac{1}{N_{pos}}L_{loc}(\hat{s}, s)  \\
  & + \frac{1}{N_{vtx}}L_{bce}(y_{lnk}, c_{lnk}) 
\end{aligned}
\end{equation}

Here, $y_{row}$ is the label of all row queries, $y_{row}^{i} = 1$ if $i$-th query is labeled as positive,, and 0 otherwise. Likewise, $y_{col}$ is the label of all column queries. $L_{bce}$ is the binary cross-entropy loss~\cite{ruby2020binary} over the predicted row and column queries scores, respectively $c_{row}$ and $c_{col}$, given by

\begin{equation}
	L(y_{r/c}, c_{r/c}) = -(ylog(p_{c})) + (1-y)log(1-p_{c}))
\end{equation}

$L_{loc}$ is the Smooth L1 regression loss~\cite{girshick2015fast} over the predicted start point geometries $\hat{s}$ and the groundtruth $s$:

\begin{equation}
	L(\hat{s}, s) = \begin{cases}
	0.5(\hat{s}-s)^{2}, if  \left| \hat{s}-s \right| < 1 \\
	\left| \hat{s}-s \right| - 0.5,  otherwise	
		   \end{cases}
\end{equation}

As for rotated angle prediction, we limit the range of rotated angles to $[-45^{\circ}, +45^{\circ}]$ and one degree represented one prediction category, the loss of rotation angle is computed as 
\begin{equation}
	L(y_{ang}, c_{ang}) = - \sum_{ang=-45}^{+45} y_{ang} log(p_{c_{ang}}) 
\end{equation}

For link classification over all vertices, we also use binary cross-entropy, given by

\begin{equation}
	L(y_{lnk}, c_{lnk}) = -(ylog(p_{c})) + (1-y)log(1-p_{c}))
\end{equation}

Notice that we only consider the loss of adjacent grids needed to merge during training. Online Hard Example Mining(OHEM)~\cite{shrivastava2016training} is applied to $L(y_{r/c}, c_{r/c})$ and $L(y_{lnk}, c_{lnk})$ for balancing positive and negative samples. 

The losses on row/column classification are normalized by $N_{r/c}$, which is the number of positive and hard negative samples. The loss on angle classification and start point regression is normalized by the number of positive samples $N_{pos}$. The loss on link classification is normalized by the number of positive and hard negative samples $N_{vtx}$

\subsection{Inference Process}
In the previous section, we introduced the structure of the model and the setting of labels. In this part, we will introduce how to get the final table structure through the output of the model. First, as shown in Fig.\ref{fig:overview}, after we put the table image into the model, we can directly obtain the output results of the Query-Based Splitting Module(QBS) and Vertex-based Merging Module(VBM). Through the output of QBS, we can get the distribution probability of horizontal and vertical lines in the table image. Set threshold $\alpha$, We can get the distribution area of the splitting line, and the connected areas represent the distribution range of a splitting line. At the same time, we define the line unit with the highest score in the range as the final splitting line. After getting the split line, we can get the distribution of basic cells in the table, and we can get the position of the vertex. That means we can get the vertex information from the output of the VBM model. According to the vertex attribute, we can merge the basic cells to get the final table structure.

\section{Experiments}
\subsection{Datasets}
Quite a number of table structure recognition datasets have been reported in recent years, and most of them can be broadly classified into two categories: standard tables and unconstrained tables. We evaluate our proposed method on the following benchmarks which contain table data of various styles in various scenes and the benchmarks are as listed below. We also provide ablation studies to verify the effects of each proposed component.

\textbf{PubTabNet\cite{zhong2020image}} PubTabNet is one of the most commonly used benchmarks for table structure recognition. It is a large-scale complicated table collection that contains 500777 training images, 9115 validating images, and 9138 testing images. This dataset contains a large amount of three-line tables with multi-row/column cells, empty cells, etc. The images of the benchmark are extracted from scientific documents.


\textbf{SynthTable}. Unlike PubTabNet which contains mostly standard tables. SynthTable covers the unconstrained table in a natural scene, requiring the table structure recognizer to have both discriminative and generative capabilities. Therefore, we also use SynthTable proposed in TIES~\cite{qasim2019rethinking}. SynthTable is a synthetic dataset that contains 1000 images for the training and 1000 images for the testing. The generated tables are harmoniously blended with the existing document background. where tables have a variety of orientations sizes and types of separators. 

\subsection{Evaluation Protocol}
In the evaluation process, we focus on the accuracy of the logical structure of the table. We use Tree-Edit-Distance-based Similarity (TEDS\cite{zhong2020image}) to evaluate the performance of our model for recognizing table logical structure. in addition to TEDs that consider both table structure and text content, we also evaluate performance on the structure TEDs metric that considers only the accuracy of table structure prediction.


\subsection{Implementation Details}
We use ResNet-18, pre-trained on ImageNet\cite{krizhevsky2012imagenet}, as the backbone, and the whole networks are then fine-tuned end-to-end using ADAM\cite{kingma2014adam} optimizer on the training sets of PubTabNet\cite{zhong2020image} and SynthTable. For fine-tuning, images are resized to $640 \times 640$ after random scaling, and the long size is resized to 640. Our model is trained for 20 epochs and the initialized learning rate is 0.0001. The batch size is set to 16. TRUST is implemented using PaddlePaddle\cite{ma2019paddlepaddle}, and we use Tesla A100 64GB GPU.

\subsection{Experimental Results}
The proposed technique is evaluated over PubTabNet and SynthTable datasets. Additionally, it is benchmarked with a number of state-of-the-art techniques such as SPLERGE\cite{tensmeyer2019deep}, TabStruct-Net\cite{raja2020table}, EDD\cite{zhong2020image}, GTE\cite{zheng2021global}, LGPMA\cite{qiao2021lgpma}, FLAG-Net\cite{liu2021show}, etc. Unlike many state-of-the-art methods that perform evaluations only at TEDs, our method also test the Structure TEDs on PubTabNet.

Tab.\ref{table:Pubtabnet} shows quantitative results on the PubTabNet dataset that contains mostly unlined tables. As the table shows, the TRUST achieves the best Structure TEDs 97.1\% and TEDs 96.2\% among all published methods for this widely studied dataset, TabStruct-Net\cite{raja2020table} has low TEDs because it cannot handle the problem of unlined tables. Our method detects row/column separators and accordingly alleviates the unlined table problem. Notice that the OCR results of PubTabNet are obtained by the public text detection method PSENet\cite{wang2019shape} and text recognition method MASTER\cite{lu2021master} for a fair comparison. The superior performance of TRUST is largely due to the proposed Query-Based Splitting Module and Vertex-based Merging Module. The individual contributions of the Query-Based Splitting Module and Vertex-based Merging Module will be discussed in the ensuing Ablation Study.

\setlength{\tabcolsep}{4pt}
\begin{table}
\begin{center}
\caption{Comparison results of logical structure recognition on PubTabNet datasets}
\label{table:Pubtabnet}
\begin{tabular}{ccc}
\hline\noalign{\smallskip}
Method & Str-TEDs & TEDs\\
\noalign{\smallskip}
\hline
\noalign{\smallskip}
EDD\cite{zhong2020image}  & -  & 88.3\% \\
TabStruct-Net\cite{raja2020table} & - & 90.1\%  \\
GTE\cite{zheng2021global} & -  &  93.0\%  \\
LGPMA\cite{qiao2021lgpma} & 96.7\% & 94.6\% \\
FLAG-Net\cite{liu2021show} & - & 95.1\% \\
\hline
Ours & 97.1\% & 96.2\% \\
\hline
\end{tabular}
\end{center}
\end{table}
\setlength{\tabcolsep}{1.4pt}

\begin{figure*}[htbp]
\centering
\includegraphics[height=10cm,width=16cm]{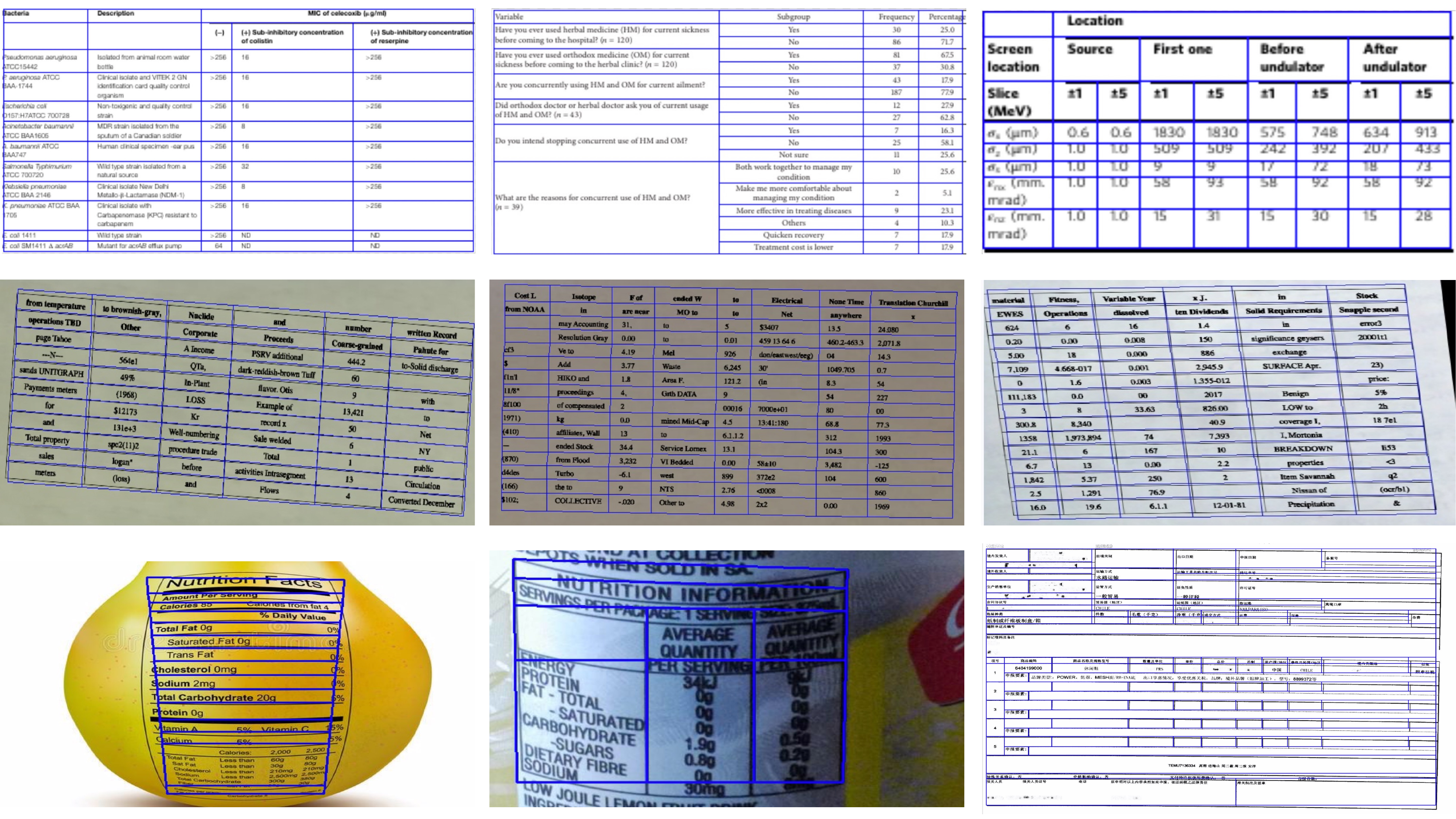}
\caption{Illustration of table structure recognition results made by TRUST: Images from first row are from PubTabNet. Images from middle row are from SynthTable. And final row is the example of bad cases. Blue lines indicate the predicted structure of tables. }
\label{fig:qualitative results}
\end{figure*}

\textbf{Results on SynthTable.} We also evaluate the SynthTable dataset proposed in TIES~\cite{qasim2019rethinking} that mainly consists of tables in diverse categories. Different from Pubtabnet, tables in SynthTable have a more diverse style such as rotation and linear perspective transformation, and a more complex background. As Tab.\ref{table:SynthTable} shows, our method achieves 99.2\%, 96.9\%, 93.6\%, and 89.2\% TEDs, respectively, outperforming the state-of-the-art methods, Both EDD and SPLERGE have a much lower TEDs because they cannot cope with tables with rotation or linear perspective transform in category 4 due to the limitation of rotation modeling. Our method models these situations through a Query-Based Splitting Module accordingly alleviates rotation and linear perspective problems. Besides, the proposed Vertex-based Merging Module explicitly merges adjacent table grids, enabling it to recognize spanning cells. This leads to up to 7.8\% and 14.3\% TEDs improvement over EDD and SPLERGE, respectively.

\textbf{Speed analysis.} We also evaluate the TRUST efficiency as shown in Tab.\ref{table:speed}. The runtime of TRUST is evaluated with NVIDIA Tesla A100 64GB. We can see that TRUST achieves 10 FPS, which is much faster than other methods such as EDD and SEM.

\setlength{\tabcolsep}{3pt}
\begin{table*}
\centering
\begin{center}
\caption{Effectiveness of Query-Based Splitting Module and Vertex-Based Merging Module on PubTabNet and SynthTable. Split: split model proposed in SPLERGE\cite{tensmeyer2019deep}, QBS: Query-Based Splitting Module, Heuristic: Heuristic Post-processing, Merge: merge model proposed in SPLERGE\cite{tensmeyer2019deep}, VBM: Vertex-Based Merging Module}
\label{table:Ablation}
\begin{tabular}{c|c|c|c|c|c||cc|}
\hline\noalign{\smallskip}
 
& \multicolumn{2}{c|}{\textbf{Splitting Model}} & \multicolumn{3}{c|}{\textbf{Merging Model}} & \multicolumn{2}{c|}{\textbf{Performance(Pubtabnet/SynthTable(C4))}} \\
\hline

 \# & \textbf{Split}\cite{tensmeyer2019deep} & \textbf{QBS} & \textbf{Heuristic\cite{tensmeyer2019deep}} & \textbf{Merge}\cite{tensmeyer2019deep} & \textbf{VBM} & Str-TEDs & TEDs\\ 
\noalign{\smallskip}
\hline
1 & \checkmark &  &  &  & \checkmark & 94.8\% / 88.2\% & 93.4\% / 85.9\%\\
2 & &  \checkmark &  \checkmark &  & & 88.3\% / 81.7\% & 85.4\% /76.7\%\\
3 & & \checkmark &  & \checkmark  &  & 96.2\% / 90.8\% & 95.3\% /86.6\%\\
4 & & \checkmark &  &  & \checkmark & 97.1\%  / 92.4\%& 96.2\% /89.2\% \\

\noalign{\smallskip}

\hline
\end{tabular}
\end{center}
\end{table*}
\setlength{\tabcolsep}{1.4pt}

\subsection{Ablation Study}
We conducted several experiments to evaluate the effectiveness of our design. These experiments mainly focus on evaluating two important modules in our TRUST: Query-based Splitting Module and Vertex-based Merging Module. Tab.\ref{table:Ablation} summarizes the results of TRUST with different settings on PubTabNet. 

\setlength{\tabcolsep}{4pt}
\begin{table}
\begin{center}
\caption{Speed analysis. TRUST is the current fastest table structure recognition method with a speed of 10 FPS. The comparisons with the previous state-of-the-arts demonstrate the efficiency of our method.}
\label{table:speed}
\begin{tabular}{cc}
\hline\noalign{\smallskip}
Method  & FPS\\
\noalign{\smallskip}
\hline
\noalign{\smallskip}
TabStruct-Net\cite{raja2020table}   & 0.77 \\
EDD\cite{zhong2020image}   & 1 \\
SEM\cite{zhang2022split}   & 1.94 \\
\hline
Ours & 10 \\
\hline
\end{tabular}
\end{center}
\end{table}
\setlength{\tabcolsep}{1.4pt}

\textbf{The Effectiveness of Query-Based Splitting Module.} We designed this module to handle the row/column separators splitting problem. To evaluate this module, we replace the Query-Based Splitting Module with the Split Model proposed in SPLERGE\cite{tensmeyer2019deep}. As Tab.\ref{table:Ablation} shows, the Split Model proposed in SPLERGE only achieves Structure TEDs 94.8\% and TEDs 93.4\%. By using the Query-Based Splitting Module, TRUST improves both Structure TEDs and TEDs by about 2.3\% and 2.8\%, respectively. The large improvement is largely due to the attention mechanism in the Query-Based Splitting Module that helps capture contextual information from long dependencies of both horizontal and vertical directions.

\textbf{The Effectiveness of Vertex-based Merging Module.} We also conducted another experiment to evaluate the Vertex-based Merging Module. We found that, if we merge the basic cells by replacing the Vertex-based Merging Module with heuristic post-processing, the Structure TEDs and TEDs drop from 97.1\% $\rightarrow$ 88.3\% and 96.2\% $\rightarrow$ 85.4\%, respectively. We further replace the Vertex-based Merging Module with the merge model proposed in SPLERGE, and the Structure TEDs and TEDs drop from 97.1\% $\rightarrow$ 96.2\% and 96.2\% $\rightarrow$ 95.3\%, respectively. This suggests that the proposed  Vertex-based Merging Module is critical to the merging results.

\textbf{The Effectiveness of Cross Feature Enhancement.} In this study, we evaluate the impact of cross-feature enhancement in the Vertex-based Merging Module by replacing them with feature enhancement from each own branch. As shown in Tab.~\ref{table:CFE}, this results in substantial performance drops, e.g., 96.2\% $\rightarrow$ 88.0\%TEDS without cross feature enhancement. suggesting that the proposed cross-feature enhancement in the Vertex-based Merging Module is an important contributor to the performance boost.

\setlength{\tabcolsep}{4pt}
\begin{table}
\begin{center}
\caption{Effectivenes of Cross Feature Enhancement in Vertex-Based Merging Module on Pubtabnet.}
\label{table:CFE}
\begin{tabular}{c|cc}
\hline\noalign{\smallskip}
 & Str-TEDs  & TEDs\\
\noalign{\smallskip}
\hline
\noalign{\smallskip}
with Cross Feature Enhancement  & 97.1\% &  96.2\% \\
w/o Cross Feature Enhancement  & 90.6\%  &   88.0\%    \\
\hline
\end{tabular}
\end{center}
\end{table}
\setlength{\tabcolsep}{1.4pt}

\subsection{Qualitative Results}  Fig.\ref{fig:qualitative results} shows a few sample images from the SynthTable dataset and the corresponding structure recognition results by TRUST. As Fig.\ref{fig:qualitative results} shows, TRUST is capable of recognizing most tables that have rotation, linear perspective transform, empty cells, spanning cells, invisible separators, etc. Its performance degrades slightly when tables appear with perspective distortion as shown in the third row in Fig.\ref{fig:qualitative results}

\section{Conclusion}
This paper presents a robust and accurate table structure recognition method using innovative encoder-decoder architecture and Transformer networks. An encoder-decoder architecture is designed that can not only reconstruct the structure of tables in arbitrary orientations but also can accurately recognize the structure of complex tables that contains spanning cells. In addition, two innovative Query-Based Splitting Module and Vertex-Based Merging Module are designed which generates feature maps with contextual information from long dependencies in a more efficient and effective way. Additionally, Transformer networks are introduced to further increase the accuracy of table structure recognition. Extensive experiments over a number of public datasets show that the proposed TRUST achieves superior performance as compared with state-of-the-art, with a remarkable faster-running speed.

\setlength{\tabcolsep}{4pt}
\begin{table*}
\begin{center}
\caption{Comparison results of logical structure recognition on SynthTable dataset, * represent models that were trained by us. C1 means standard tables with visible lines; C2 means standard tables without invisible lines; C3 means standard tables with spanning cells; C4 means unconstrained tables with rotation and linear perspective transform. }
\label{table:SynthTable}
\begin{tabular}{c|cc|cc|cc|cc}
\hline\noalign{\smallskip}\
& \multicolumn{2}{c|}{\textbf{C1}} & \multicolumn{2}{c|}{\textbf{C2}} & \multicolumn{2}{c|}{\textbf{C3}} & \multicolumn{2}{c}{\textbf{C4}}\\
\hline
Method  & Str-TEDs & TEDs & Str-TEDs & TEDs & Str-TEDs & TEDs & Str-TEDs & TEDs \\
\noalign{\smallskip}
\hline
\noalign{\smallskip}
$EDD^{*}$\cite{long2021parsing} & 97.8\% &  96.0\% & 98.0\% & 93.4\%   & 96.1\%  & 93.2\%  & 89.9\%  & 81.4\% \\
$SPLERGE^{*}$\cite{zhong2020image}  & 97.8\%  &  97.0\%   &  94.4\%   & 91.6\%  &  95.5\%  & 92.1\% &  85.6\%   & 74.9\%\\
TRUST w/o\cite{tensmeyer2019deep} & 99.6\% &  99.0\%  & 98.0\%    & 97.0\%  & 96.1\%    & 93.6\%  & 90.7\%    & 81.0\% \\
\hline
TRUST  & 99.7\% & 99.2\% &  98.1\% &  96.9\% &  96.0\% &  93.6\% &  92.4\% &  89.2\% \\
\hline
\end{tabular}
\end{center}
\end{table*}
\setlength{\tabcolsep}{1.4pt}

\bibliographystyle{IEEEtran}
\bibliography{bare_jrnl_new_sample4.bbl}

\end{document}